\documentclass[runningheads]{llncs}
\usepackage[T1]{fontenc}
\usepackage{graphicx,verbatim}
\usepackage{amsmath} 
\usepackage{amssymb}
\usepackage{graphicx}
\usepackage{subcaption}
\usepackage{adjustbox}
\usepackage{booktabs}
\usepackage{multirow}

\begin{document}
\title{DualPrompt-MedCap: A Dual-Prompt Enhanced Approach for Medical Image Captioning}
\titlerunning{DualPrompt-MedCap}
%

\author{Yining Zhao \and Ali Braytee \and Mukesh Prasad}

\authorrunning{Y. Zhao et al.}

\institute{University of Technology Sydney, Sydney, Australia \\
\email{\{Yining.Zhao@student, Ali.Braytee, Mukesh.Prasad\}@uts.edu.au}}
    
\maketitle              
\begin{abstract}

Medical image captioning via vision-language models has shown promising potential for clinical diagnosis assistance. However, generating contextually relevant descriptions with accurate modality recognition remains challenging. We present DualPrompt-MedCap, a novel dual-prompt enhancement framework that augments Large Vision-Language Models (LVLMs) through two specialized components: (1) a modality-aware prompt derived from a semi-supervised classification model pre-trained on medical question-answer pairs, and (2) a question-guided prompt leveraging biomedical language model embeddings. To address the lack of captioning ground truth, we also propose an evaluation framework that jointly considers spatial-semantic relevance and medical narrative quality. Experiments on multiple medical datasets demonstrate that DualPrompt-MedCap outperforms the baseline BLIP-3 by achieving a 22\% improvement in modality recognition accuracy while generating more comprehensive and question-aligned descriptions. Our method enables the generation of clinically accurate reports that can serve as medical experts' prior knowledge and automatic annotations for downstream vision-language tasks.


\end{abstract}
\section{Introduction}
Medical image understanding plays a crucial role in clinical diagnosis, yet the interpretation and reporting of such images still heavily rely on experienced pathologists \cite{farahani2015whole}, especially challenging in regions with limited expertise \cite{niazi2019digital}. Vision-language models (VLMs) have shown promise in automating medical image captioning \cite{li2018hybrid,wang2018tienet}, but face three critical limitations: (1) unreliable modality recognition, where models struggle to distinguish between imaging techniques \cite{jing2018automatic}; (2) generic prompts lacking medical specificity; and (3) reliance on large annotated datasets, which are scarce due to privacy concerns \cite{liu2019chest}. Early captioning approaches adapted generic models like Show and Tell \cite{vinyals2015show} to specialized architectures such as R2Gen \cite{chen2020generating}, but typically overlooked modality awareness—crucial for diagnostic accuracy. While prompt-based methods \cite{radford2021learning} and BLIP-based models \cite{li2023blip} show promise, they lack medical-specific constraints. The scarcity of labeled medical data compounds these challenges, with semi-supervised techniques like FixMatch \cite{sohn2020fixmatch} showing potential but remaining underexplored in medical contexts \cite{lu2021data}. Additionally, evaluation metrics like BLEU \cite{papineni2002bleu} and domain-specific frameworks such as CheXpert \cite{irvin2019chexpert} and RadGraph \cite{jain2021radgraph} rely on ground truth references and often fail to capture true clinical relevance.

To address these limitations, we propose DualPrompt-MedCap, a question-guided medical image captioning framework with three key innovations: (1) a semi-supervised learning approach with novel Medical Modality Attention mechanisms that significantly improves modality recognition accuracy with limited labeled data; (2) a dual-prompt strategy that integrates modality-aware prompts with question-guided clinical focus to generate contextually appropriate descriptions; and (3) a ground truth-independent evaluation framework that jointly assesses image relevance, question alignment, and adherence to radiological standards. Our experimental results demonstrate that this approach substantially outperforms baseline methods in generating clinically accurate and question-relevant medical image descriptions.

\section{DualPrompt-MedCap framework}

\begin{figure}[t]
    \centering
    \includegraphics[width=\textwidth]{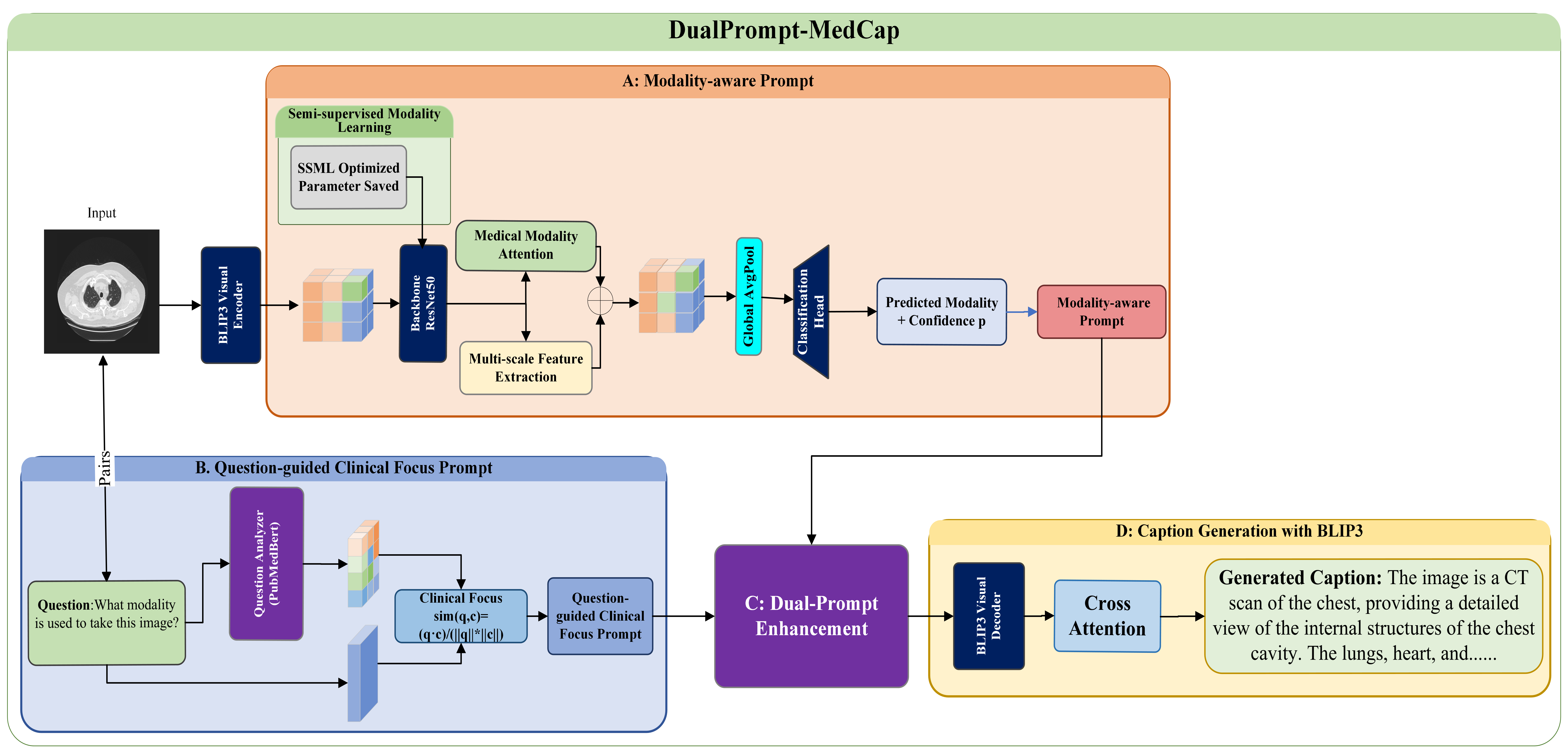}
    \caption{Overview of the proposed DualPrompt-MedCap framework.}
    \label{fig:dualprompt}
\end{figure}

We aim to build a question-guided medical image captioning system with enhanced modality recognition capabilities. It involves two challenges: (1) accurately identifying medical imaging modalities with limited labeled data, and (2) generating descriptions that integrate both modality-specific information and question-relevant clinical context. Vision-language models like BLIP3 often misclassify medical imaging modalities and use generic prompts. To address these challenges, we propose DualPrompt-MedCap, a framework leveraging semi-supervised learning and specialized attention mechanisms. Our framework (Fig.~\ref{fig:dualprompt}) consists of four main components: a modality-aware prompt generation module (A), a question-guided clinical focus analysis system (B), a dual-prompt enhancement mechanism (C), and a caption generation module powered by BLIP3 (D). By separating modality recognition from content description, DualPrompt-MedCap eliminates a common source of errors while focusing on diagnostically relevant features.

\subsection{Modality-aware Prompt}

\begin{figure}[t]
    \centering
    \includegraphics[width=\textwidth]{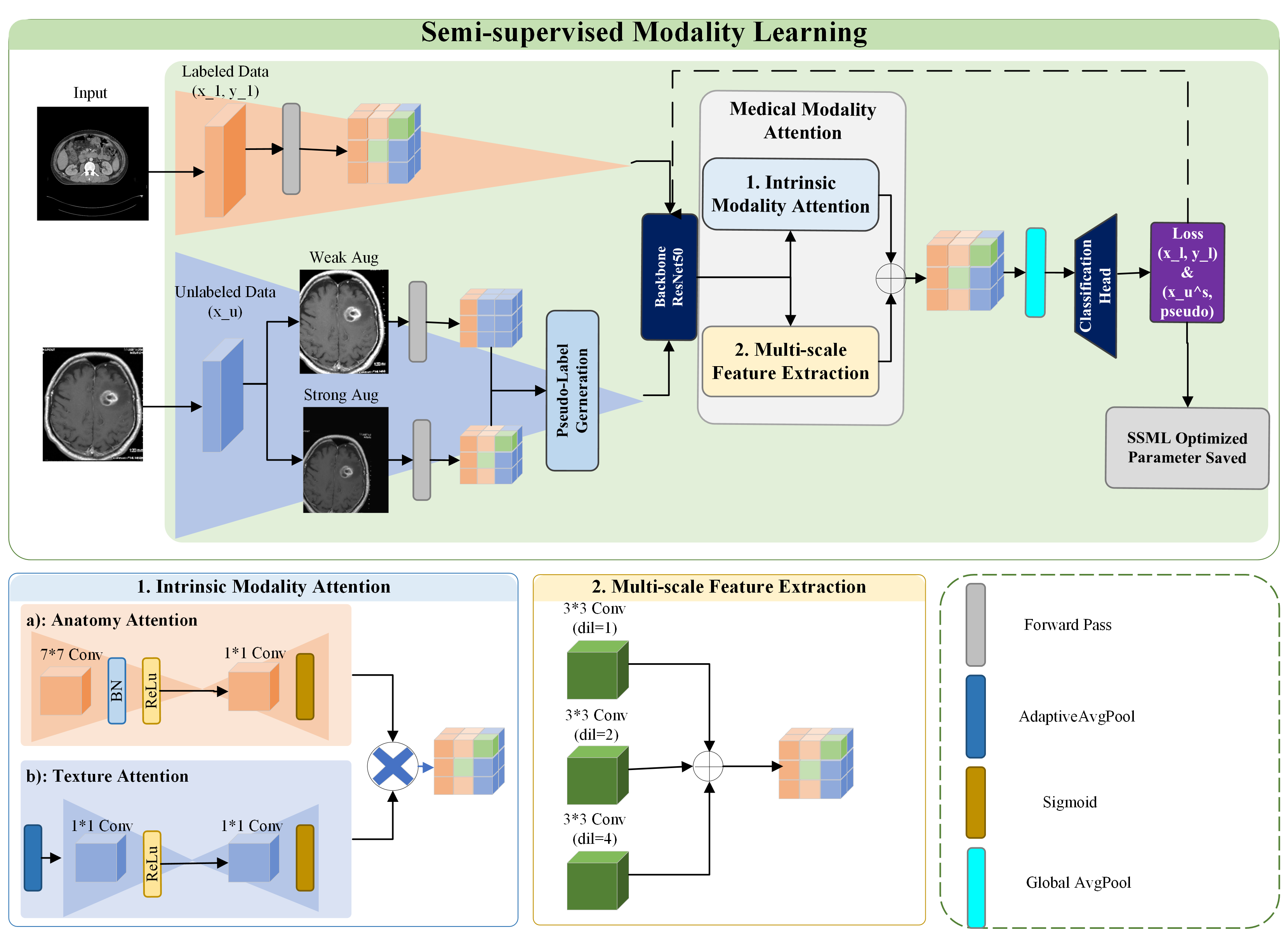}
    \caption{Semi-supervised modality learning module.}
    \label{fig:SSLframework}
\end{figure}

The Modality-aware Prompt component leverages a pre-trained modality classifier for accurate imaging technique identification (CT, MRI, or X-ray). This classifier processes medical images through BLIP3's visual encoder followed by our enhanced backbone with Medical Modality Attention, incorporating the predicted modality into a structured prompt template.

\subsubsection{Semi-supervised Modality Learning}
We propose semi-supervised modality learning (Fig.~\ref{fig:SSLframework}) by extending FixMatch~\cite{sohn2020fixmatch} to address limited labeled modality data. The framework processes labeled data $(x_l, y_l)$ through a supervised pathway and unlabeled data $(x_u)$ through weak-to-strong consistency training. For labeled data, we compute supervised loss using modality-specific class weights. For unlabeled data, weak augmentation generates pseudo-labels that guide learning from strongly augmented versions, incorporating only pseudo-labels exceeding a confidence threshold $\tau$.

\subsubsection{Medical Modality Attention}
We introduce a Medical Modality Attention mechanism that adaptively emphasizes modality-specific visual cues through two components: Intrinsic Modality Attention and Multi-scale Feature Extraction. The feature transformation process is:
\begin{equation}
F_{\text{out}} = x \odot A_{\text{intr}}(x) + M_{\text{scale}}(x)
\end{equation}
where $A_{\text{intr}}$ represents Intrinsic Modality Attention, $M_{\text{scale}}$ represents Multi-scale Feature Extraction, and $\odot$ denotes element-wise multiplication.

\subsubsection{Intrinsic Modality Attention}
The Intrinsic Modality Attention module captures anatomical structures and modality-specific imaging patterns:
\begin{equation}
A_{\text{intr}}(x) = A_{\text{anatomy}}(x) \odot A_{\text{texture}}(x)
\end{equation}

\textbf{a) Anatomy Attention:} Implemented via $7 \times 7$ convolutions to enhance organ and tissue boundary detection, critical for distinguishing anatomical structures across modalities. Here, $A_{\text{anatomy}}(x)$ employs batch normalization and ReLU activation.

\textbf{b) Texture Attention:} Captures unique contrast patterns of each modality through channel-wise operations, where $A_{\text{texture}}(x)$ applies adaptive average pooling followed by two $1 \times 1$ convolutions with a dimension reduction and expansion strategy.

\subsubsection{Multi-scale Feature Extraction}
Processes features at different anatomical scales through dilated convolutions:
\begin{equation}
M_{\text{scale}}(x) = W_{\text{adj}}\left(\bigoplus_{d\in\{1,2,4\}} \text{Conv}_d(x)\right)
\end{equation}
where $\text{Conv}_d$ represents dilated convolutions with dilation rates of 1, 2, and 4, $W_{\text{adj}}$ is $1\times1$ convolutions for channel adjustment, and $\bigoplus$ represents channel-wise concatenation.

Our semi-supervised strategy includes two key optimizations:

\textbf{Modality-specific Augmentation:} We address VLMs' tendency to misclassify MRI as CT through targeted augmentation strategies. For MRI, we apply enhanced geometric transformations and controlled intensity adjustments to preserve characteristic tissue contrast while increasing pattern diversity.

\textbf{Semi-supervised Loss:} We adapt FixMatch with a modified loss function for medical imaging:
\begin{equation}
\mathcal{L} = \sum_{(x_l,y_l)} w_{y_l}\text{CE}(f(x_l), y_l) + \lambda \sum_{x_u} \mathbb{I}(p > \tau) \cdot \text{CE}(f(x_u^s), \hat{y}_u)
\end{equation}
where $w_{y_l}$ are modality-specific class weights, $\hat{y}_u$ are pseudo-labels from weakly augmented samples, and $\tau$ (0.95) filters unreliable predictions.

\subsubsection{Modality Prediction and Prompt Generation}
During application, our framework loads parameters from the semi-supervised learning stage. The input image is processed through BLIP3's visual encoder and our enhanced backbone to predict modality and construct a modality-aware prompt, providing crucial technical context for caption generation.

\subsection{Question-guided Clinical Focus Prompt}
The Question-guided Clinical Focus Prompt component employs a Question Analyzer built on PubMedBERT~\cite{gu2021domain} to extract semantic information from clinical queries. This component identifies both question types and relevant clinical concepts. The Question Analyzer computes clinical focus using cosine similarity between question embeddings and predefined clinical concept embeddings:
\begin{equation}
\text{sim}(q, c) = \frac{q \cdot c}{\|q\| \cdot \|c\|}
\end{equation}
where $q$ represents question embedding and $c$ denotes embeddings of clinical concepts across categories including anatomy, pathology, location, and comparison.

The analyzer maintains comprehensive medical concept dictionaries covering anatomical structures, pathological findings, spatial locations, and comparison terms. Beyond semantic similarity, it conducts syntactic and contextual analysis to classify question types and extract relevant terms. For example, with "Which side of the lung is abnormal?", it identifies "lung" as anatomical focus, "abnormal" as pathology-related, and classifies it as a location-type question.

\subsection{Dual-Prompt Enhancement}
The Dual-Prompt Enhancement integrates the modality-aware prompt and question-guided clinical focus prompt into a unified guidance signal. This creates a comprehensive context capturing both imaging technique characteristics and clinical query focus. For instance, examining a chest CT with a question about lung abnormalities would create a prompt specifying both "CT scan of the chest" and "examining for lung abnormalities". By fusing these complementary prompts, our approach addresses the limitations of generic prompting methods, focusing caption generation on clinically relevant aspects and improving both accuracy and utility for medical professionals.

\subsection{Caption Generation with BLIP3}
The caption generation module powered by BLIP3~\cite{xue2024xgen} transforms integrated dual prompts and visual features into coherent, clinically relevant descriptions. BLIP3's visual decoder processes the input image's features, combined with dual-prompt information through cross-attention, enabling selective focus on image regions most relevant to both modality context and clinical question.

\subsection{A Ground Truth-Independent Evaluation Scores}
To address the limitations of reference-dependent metrics, we propose a novel ground truth-independent evaluation framework integrating multimodal relevance assessment with medical quality evaluation. Since our approach focuses on question-guided descriptions, we evaluate both image-caption alignment and question-response correlation using BiomedCLIP~\cite{zhang2023biomedclip} to compute cosine similarity between image and caption embeddings ($S_{\text{image-text}}$) and between question and caption embeddings ($S_{\text{question-text}}$). The overall relevance score combines these similarities:
\begin{equation}
S_{\text{relevance}} = \alpha_1 S_{\text{image-text}} + \alpha_2 S_{\text{question-text}}
\end{equation}
This dual-similarity approach ensures captions remain visually grounded while addressing the clinical inquiry. For medical quality evaluation, follow established standards in radiological reporting~\cite{schwartz2011improving}, we evaluate three critical aspects:
\begin{equation}
S_{\text{quality}} = \beta_1 S_{\text{medical}} + \beta_2 S_{\text{clinical}} + \beta_3 S_{\text{structure}}
\end{equation}

Our framework assesses medical terminology ($S_{\text{medical}}$) using UMLS~\cite{bodenreider2004unified} through spaCy's scispacy\_linker~\cite{neumann2019scispacy}, measuring density and diversity of standard terms. We evaluate clinical correctness ($S_{\text{clinical}}$) by assessing anatomical localization, finding identification, and measurement precision. Additionally, we evaluate report structure ($S_{\text{structure}}$) to ensure consistency with radiological reporting conventions. Finally, We integrate relevance and medical quality into a comprehensive score:
\begin{equation}
S_{\text{final}} = \gamma_1 S_{\text{relevance}} + \gamma_2 S_{\text{quality}}
\end{equation}
This integration prioritizes clinical accuracy and contextual appropriateness in multi-criteria evaluation~\cite{reiter2009investigation,liu2019clinically}.

\section{Experiments}
\subsection{Datasets and Implementation Details}
We conduct experiments on two VQA medical datasets: RAD~\cite{lau2018dataset} and SLAKE~\cite{liu2021slake}. For modality learning, we extract labeled data from RAD by identifying modality keywords (``mri'', ``ct'', ``xray'') in questions or answers, supporting our semi-supervised approach. SLAKE provides complete modality ground truth and paired images, ideal for caption generation evaluation.


\subsection{Experiment Settings}
Our semi-supervised modality learning module is implemented with PyTorch using ResNet50 pre-trained on ImageNet as the backbone network. For MRI images, we apply stronger geometric transformations (i.e., $\pm15^\circ$ rotation, 15\% translation) to address their unique characteristics and mitigate misclassification with CT scans. The FixMatch training strategy uses a confidence threshold of 0.95 for pseudo-labeling. For caption generation, we leverage BLIP3 as the base vision-language model and PubMedBERT for the Question Analyzer. In our evaluation framework, we balance image-text and question-text similarities (i.e., $\alpha_1 = \alpha_2 = 0.25$) to ensure both visual alignment and clinical question matching contribute equally. We also equally weight medical terminology, clinical correctness, and report structure (i.e., $\beta_1 = \beta_2 = \beta_3 = \frac{1}{3}$) in quality assessment, reflecting their interdependent nature in radiological reporting. Our integrated evaluation gives equal importance to relevance and quality scores ($\gamma_1 = \gamma_2 = 1$).

\section{Results}
\begin{table}
\scriptsize
\vspace{-0.4cm}  
\caption{Modality classification accuracy on SLAKE dataset.}\label{tab:modality_classification}
\begin{tabular}{|l|c|c|c|c|}
\toprule
\textbf{Modality} & \textbf{Semi-supervised Modality} & \textbf{BLIP3\cite{xue2024xgen}} & \textbf{FixMatch\cite{sohn2020fixmatch}} & \textbf{Total Samples}\\
\midrule
MRI & \textbf{92.11\%} & 11.84\% & 73.30\% & 472\\
CT & \textbf{100.00\%} & 92.37\% & 94.49\% & 361\\
X-RAY & \textbf{100.00\%} & \textbf{100.00\%} & \textbf{100.00\%} & 228\\
\midrule
Average Accuracy & \textbf{98.30\%} & 77.66\% & 86.25\% & 1061\\
\bottomrule
\end{tabular}
\vspace{-0.4cm}  
\end{table}

\subsection{Modality Classification Results}
We first demonstrate the ability of our proposed semi-supervised modality learning (FixMatch+Attention) to predict modality in datasets where this label does not exist, to be included in the prompt. Table~\ref{tab:modality_classification} presents modality classification results on the SLAKE dataset. Our DualPrompt-MedCap framework significantly outperforms baseline methods, particularly in MRI recognition (92.11\% vs BLIP3's 11.84\%). The substantial performance gap (98.30\% vs 77.66\% average accuracy) confirms that specialized modality recognition is essential for generating clinically useful medical image captions.

\subsection{Medical Captioning Evaluation Results}
\begin{table}[t]
\scriptsize
\vspace{-0.4cm}  
\caption{Evaluation results on SLAKE and RAD datasets}\label{tab:eval_results}
\begin{tabular}{|l|c|c|c|c|c|c|c|c|}
\toprule
\multirow{2}{*}{\textbf{Metric}} & \multicolumn{4}{c|}{\textbf{SLAKE Dataset}} & \multicolumn{4}{c|}{\textbf{RAD Dataset}} \\
\cmidrule(lr){2-5} \cmidrule(lr){6-9}
 & \textbf{Ours} & \textbf{BLIP3\cite{xue2024xgen}} & \textbf{Tag2Text\cite{huang2023tag2text}} & \textbf{BLIP2\cite{li2023blip}} & \textbf{Ours} & \textbf{BLIP3} & \textbf{Tag2Text} & \textbf{BLIP2} \\ 
\midrule
Final Score & \textbf{0.5396} & 0.4716 & 0.2956 & 0.3273 & \textbf{0.5337} & 0.4494 & 0.3332 & 0.3318 \\
\midrule
\multicolumn{9}{|c|}{\normalsize Relevance Assessment} \\
\midrule
Image Similarity & \textbf{0.3841} & 0.3521 & 0.3070 & 0.3724 & \textbf{0.3608} & 0.3171 & 0.3249 & 0.3173 \\
Question Similarity & \textbf{0.5694} & 0.5323 & 0.5616 & 0.5585 & \textbf{0.5929} & 0.5013 & 0.5777 & 0.5728 \\
\midrule
\multicolumn{9}{|c|}{\normalsize Medical Quality Evaluation} \\
\midrule
Medical Quality & 0.3514 & 0.3545 & \textbf{0.4593} & 0.4476 & 0.3854 & 0.3717 & 0.4741 & \textbf{0.4862} \\
Clinical Accuracy & \textbf{0.5833} & 0.3348 & 0.0272 & 0.0177 & \textbf{0.5563} & 0.3207 & 0.0114 & 0.0915 \\
Structure & \textbf{0.8737} & 0.8398 & 0.1482 & 0.1108 & \textbf{0.8302} & 0.8035 & 0.2492 & 0.1588 \\
\bottomrule
\end{tabular}
\vspace{-0.4cm}  
\end{table}

Table~\ref{tab:eval_results} shows evaluation results on SLAKE and RAD datasets. DualPrompt-MedCap achieves the highest final scores on both datasets (0.5396 and 0.5337), outperforming all baseline models. The most significant improvements are in question similarity (0.5694/0.5929) and clinical accuracy (0.5833/0.5563), demonstrating our model's ability to generate captions aligned with clinical queries while maintaining medical relevance. Our approach also excels in structural consistency (0.8737/0.8302), in contrast to Tag2Text and BLIP2, which show poor clinical accuracy (<0.1) despite competitive medical terminology scores. These results validate that our dual-prompt approach significantly enhances clinical utility across different medical datasets.

\subsection{Qualitative Analysis of DualPrompt-MedCap}
\begin{figure}[htbp]
\vspace{-0.3cm}  
\centering
\includegraphics[width=1\textwidth]{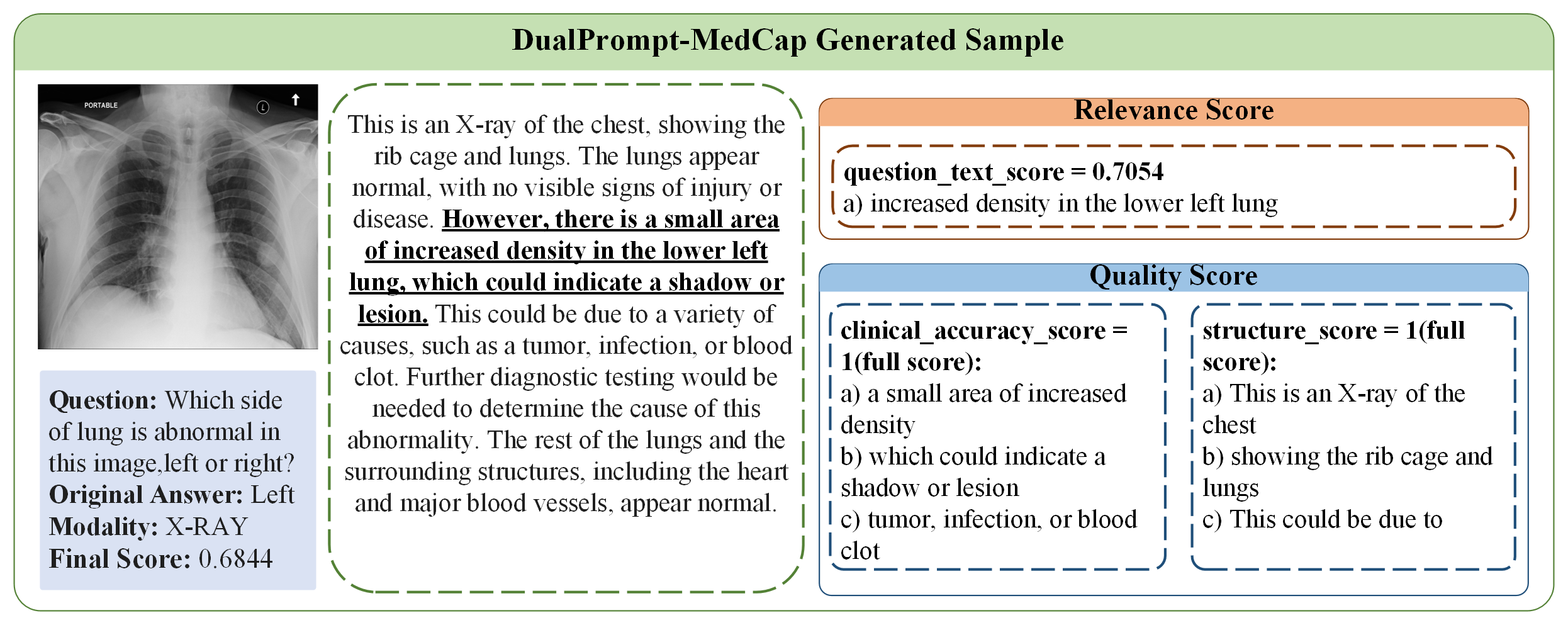}
\caption{DualPrompt-MedCap caption analysis showing relevance and quality.}
\label{fig:medcap_sample}
\vspace{-0.4cm} 
\end{figure}
Fig.~\ref{fig:medcap_sample} shows an analysis of a caption generated by DualPrompt-MedCap (final score: 0.6844). The caption directly addresses the query about lung abnormality by identifying "increased density in the lower left lung," demonstrating strong question relevance (0.7054). The caption received perfect scores in clinical accuracy and report structure (1.0000 each), demonstrating how DualPrompt-MedCap integrates modality awareness and question guidance to generate both technically accurate and clinically meaningful descriptions.
\begin{figure}[htbp]
\centering
\includegraphics[width=1\textwidth]{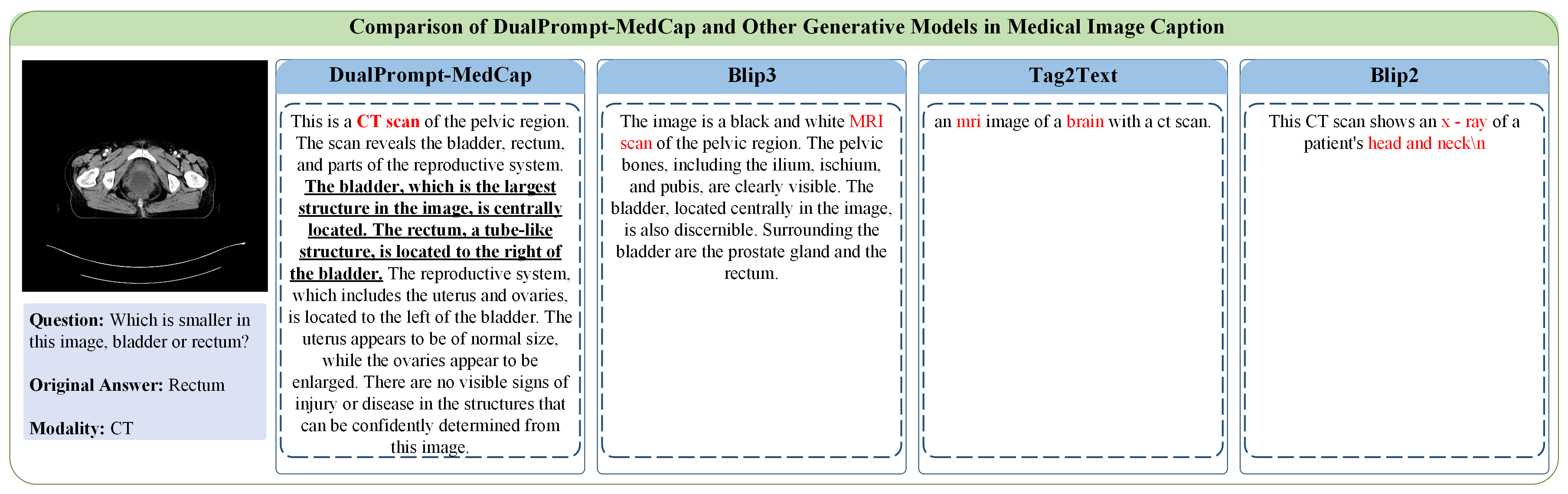}
\caption{DualPrompt-MedCap with other models in medical image captioning.}
\label{fig:dualprompt_comparison}
\vspace{-0.8cm} 
\end{figure}
Fig.~\ref{fig:dualprompt_comparison} compares captions from different models. While DualPrompt-MedCap accurately identifies the CT scan and addresses the clinical question, other models either misidentify the modality (BLIP3, Tag2Text, and BLIP2 all incorrectly label the imaging type) or fail to address the clinical query. DualPrompt-MedCap's caption includes precise anatomical descriptions, demonstrating how our dual-prompt strategy effectively integrates modality awareness and question relevance for clinically useful medical image descriptions.

\section{Conclusion}
In this paper, we proposed DualPrompt-MedCap, a framework for question-guided medical image captioning with enhanced modality recognition. By combining semi-supervised learning and a medical modality attention mechanism, we improved modality recognition accuracy. Our approach ensures both technical accuracy and clinical relevance, producing captions that meet radiological standards. Experimental results on VQA datasets highlight its superior performance, and potential to assist medical professionals, reduce radiologist workload, and improve diagnostic workflows, especially in resource-limited settings.

\end{document}